\PassOptionsToPackage{unicode}{hyperref}
\PassOptionsToPackage{hyphens}{url}
\PassOptionsToPackage{dvipsnames,svgnames,x11names}{xcolor}
\documentclass[
]{article}
\usepackage{amsmath,amssymb}
\usepackage[left=2cm,right=2cm]{geometry}
\usepackage{lmodern}
\usepackage{iftex}
\ifPDFTeX
  \usepackage[T1]{fontenc}
  \usepackage[utf8]{inputenc}
  \usepackage{textcomp} 
\else 
  \usepackage{unicode-math}
  \defaultfontfeatures{Scale=MatchLowercase}
  \defaultfontfeatures[\rmfamily]{Ligatures=TeX,Scale=1}
\fi
\IfFileExists{upquote.sty}{\usepackage{upquote}}{}
\IfFileExists{microtype.sty}{
  \usepackage[]{microtype}
  \UseMicrotypeSet[protrusion]{basicmath} 
}{}
\makeatletter
\@ifundefined{KOMAClassName}{
  \IfFileExists{parskip.sty}{%
    \usepackage{parskip}
  }{
    \setlength{\parindent}{0pt}
    \setlength{\parskip}{6pt plus 2pt minus 1pt}}
}{
  \KOMAoptions{parskip=half}}
\makeatother
\usepackage{xcolor}
\usepackage{graphicx}
\makeatletter
\def\maxwidth{\ifdim\Gin@nat@width>\linewidth\linewidth\else\Gin@nat@width\fi}
\def\maxheight{\ifdim\Gin@nat@height>\textheight\textheight\else\Gin@nat@height\fi}
\makeatother
\setkeys{Gin}{width=\maxwidth,height=\maxheight,keepaspectratio}
\makeatletter
\def\fps@figure{htbp}
\makeatother
\newlength{\cslhangindent}
\newlength{\csllabelwidth}
\newlength{\cslentryspacingunit} 
\newenvironment{CSLReferences}[2] 
 {
  \setlength{\parindent}{0pt}
  \ifodd #1
  \let\oldpar\par
  \def\par{\hangindent=\cslhangindent\oldpar}
  \fi
  \setlength{\parskip}{#2\cslentryspacingunit}
 }%
 {}
\usepackage{calc}

\ifLuaTeX
\usepackage[bidi=basic]{babel}
\else
\usepackage[bidi=default]{babel}
\fi
\babelprovide[main,import]{american}

\def\languageshorthands#1{}
\ifLuaTeX
  \usepackage{selnolig}  
\fi
\IfFileExists{bookmark.sty}{\usepackage{bookmark}}{\usepackage{hyperref}}
\IfFileExists{xurl.sty}{\usepackage{xurl}}{} 
\hypersetup{
  pdftitle={MLMOD: Machine Learning Methods for Data-Driven Modeling in
LAMMPS},
  pdfauthor={Paul J. Atzberger},
  pdflang={en-US},
  colorlinks=true,
  linkcolor={Maroon},
  filecolor={Maroon},
  citecolor={Blue},
  urlcolor={Blue},
  pdfcreator={LaTeX via pandoc}}

\title{MLMOD: Machine Learning Methods for Data-Driven Modeling in
LAMMPS}

\usepackage[affil-it]{authblk}
\usepackage{orcidlink}
\author[1%
  ]{Paul J. Atzberger%
    \,\orcidlink{0000-0001-6806-8069}\,%
    }

\affil[1]{Paul J. Atzberger, Professor, University of California Santa
Barbara}
\date{}

\begin{document}
\maketitle

\hypertarget{summary}{%
\section{Summary}\label{summary}}

\texttt{MLMOD} is a software package for incorporating machine learning
approaches and models into simulations of microscale mechanics and
molecular dynamics in LAMMPS. Recent machine learning approaches provide
promising data-driven approaches for learning representations for system
behaviors from experimental data and high fidelity simulations. The
package faciliates learning and using data-driven models for (i)
dynamics of the system at larger spatial-temporal scales (ii)
interactions between system components, (iii) features yielding coarser
degrees of freedom, and (iv) features for new quantities of interest
characterizing system behaviors. \texttt{MLMOD} provides hooks in LAMMPS
for (i) modeling dynamics and time-step integration, (ii) modeling
interactions, and (iii) computing quantities of interest characterizing
system states. The package allows for use of machine learning methods
with general model classes including Neural Networks, Gaussian Process
Regression, Kernel Models, and other approaches. Here we discuss our
prototype C++/Python package, aims, and example usage. The package is
integrated currently with the mesocale and molecular dynamics simulation
package LAMMPS and PyTorch. The source code for this initial version 1.0
of \texttt{MLMOD} has been archived to Zenodo
(\protect\hyperlink{ref-zenodo}{P. J. Atzberger, 2023}). For related
papers, examples, updates, and additional information see
\url{https://github.com/atzberg/mlmod} and \url{http://atzberger.org/}.

\hypertarget{statement-of-need}{%
\section{Statement of Need}\label{statement-of-need}}

A practical challenge in using machine learning methods for simulations
is the efforts required to incorporate learned system features to
augment existing models and simulation methods. Our package
\texttt{MLMOD} aims to address this aspect of data-driven modeling by
providing a general interface for incorporating ML models using
standardized representations and by leveraging existing simulation
frameworks such as LAMMPS
(\protect\hyperlink{ref-Plimpton:2022}{Thompson et al., 2022}). Our
\texttt{MLMOD} package provides hooks which are triggered during key
parts of simulation calculations. In this way standard machine learning
frameworks can be used to train ML models, such as PyTorch
(\protect\hyperlink{ref-Paszke:2019}{Paszke et al., 2019}) and
TensorFlow (\protect\hyperlink{ref-Abadi:2015}{Abadi et al., 2015}),
with the resulting models more amenable to being translated into
practical simulations. The models obtained from learning can be
accommodated in many forms, including Deep Neural Networks (DNNs)
(\protect\hyperlink{ref-Goodfellow:2016}{Goodfellow et al., 2016}),
Kernel Regression Models (KRM)
(\protect\hyperlink{ref-Scholkopf:2001}{Scholkopf \& Smola, 2001}),
Gaussian Process Regression (GPR)
(\protect\hyperlink{ref-Rasmussen:2004}{Rasmussen, 2004}), and others
(\protect\hyperlink{ref-Hastie:2001}{Hastie et al., 2001}).

\hypertarget{data-driven-modeling}{%
\section{Data-Driven Modeling}\label{data-driven-modeling}}

Recent advances in machine learning, optimization, and available
computational resources are presenting new opportunities for data-driven
modeling and simulation in the natural sciences and engineering.
Empirical successes in deep learning suggest promising non-linear
techniques for learning representations for system behaviors and other
underlying features (\protect\hyperlink{ref-Goodfellow:2016}{Goodfellow
et al., 2016}; \protect\hyperlink{ref-Hinton:2006}{Hinton \&
Salakhutdinov, 2006}). Many previous deep learning methods have been
developed for problems motivated by image analysis and natural language
processing. However, scientific computations and associated dynamical
systems present a unique set of challenges for developing and employing
recent machine learning approaches
(\protect\hyperlink{ref-Atzberger:2018}{P. J. Atzberger, 2018};
\protect\hyperlink{ref-Brunton:2016}{Brunton et al., 2016};
\protect\hyperlink{ref-Schmidt:2009}{Schmidt \& Lipson, 2009}).

In scientific and engineering applications there are often important
constraints arising from physical principles required to obtain
plausible models and there is a need for results to be more
interpretable. In large-scale scientific computations, bottom-up
modeling efforts aim to start as close as possible to first principles
and perform computations to obtain insights into larger-scale emergent
behaviors. Examples include the rheological responses of soft materials
and complex fluids from microstructure interactions
(\protect\hyperlink{ref-Atzberger:2013}{Paul J. Atzberger, 2013};
\protect\hyperlink{ref-Bird:1987}{Bird, 1987};
\protect\hyperlink{ref-Kimura:2009}{Kimura, 2009};
\protect\hyperlink{ref-Lubensky:1997}{Lubensky, 1997}), molecular
dynamics modeling of protein structures and functional domains from
atomic level interactions (\protect\hyperlink{ref-Karplus:1983}{Brooks
et al., 1983}; \protect\hyperlink{ref-Karplus:2002}{Karplus \& McCammon,
2002}; \protect\hyperlink{ref-Mccammon:1988}{McCammon \& Harvey, 1988};
\protect\hyperlink{ref-Plimpton:2022}{Thompson et al., 2022}), and
prediction of weather and climate phenomena from detailed physical
models, sensor data, and other measurements
(\protect\hyperlink{ref-Bauer:2015}{Bauer et al., 2015};
\protect\hyperlink{ref-Richardson:2007}{Richardson, 2007}). Obtaining
observables and quantities of interest (QoI) from simulations of such
high fidelity detailed models can involve significant computational
resources (\protect\hyperlink{ref-Giessen:2020}{Giessen et al., 2020};
\protect\hyperlink{ref-Lusk:2011}{Lusk \& Mattsson, 2011};
\protect\hyperlink{ref-Murr:2016}{Murr, 2016};
\protect\hyperlink{ref-Pan:2021}{Pan, 2021};
\protect\hyperlink{ref-Sanbonmatsu:2007}{Sanbonmatsu \& Tung, 2007};
\protect\hyperlink{ref-Washington:2009}{Washington et al., 2009}).
Data-driven learning methods present opportunities to formulate more
simplified models, provide model flexibility to accommodate subtle
effects, or make predictions which are less computationally expensive.

Data-driven modeling can take many forms. As a specific motivation for
the package and our initial implementations, we discuss a specific case
in detail, but our package also can be used more broadly. In particular,
we consider detailed molecular dynamics simulations of large spherical
colloidal particles within a bath of much smaller solvent particles. A
common problem is to infer interaction laws between the colloidal
particles given the surrounding environment arising from the type of
solution, charge, and other physical conditions. There is extensive
theoretical literature on colloidal interactions and approximate models
(\protect\hyperlink{ref-Derjaguin:1941}{Derjaguin, 1941};
\protect\hyperlink{ref-Doi:2013}{Doi, 2013};
\protect\hyperlink{ref-Jones:2002}{Jones et al., 2002}). While analytic
approaches have had success, there are many settings where challenges
remain which limit the accuracy
(\protect\hyperlink{ref-Jones:2002}{Jones et al., 2002};
\protect\hyperlink{ref-Atzberger:2018b}{Sidhu et al., 2018}).
Computational modeling and simulation provides opportunities for
capturing phenomena in more physical detail and with better
understanding of contributing effects.

While simulations of colloids including the solvent and other
environmental factors can be used for making predictions, such
computations can be expensive given the many degrees of freedom and
small time-scales of solvent-solvent interactions. Colloid
coarse-grained models are sought which utilize separation in scales,
such as the contrast in size with the solvent and dynamical time-scales.
In these circumstances, coarse-grained models aim to capture the
effective colloidal interactions and their dynamics.

\begin{figure}
\centering
\includegraphics[width=0.75\textwidth,height=\textheight]{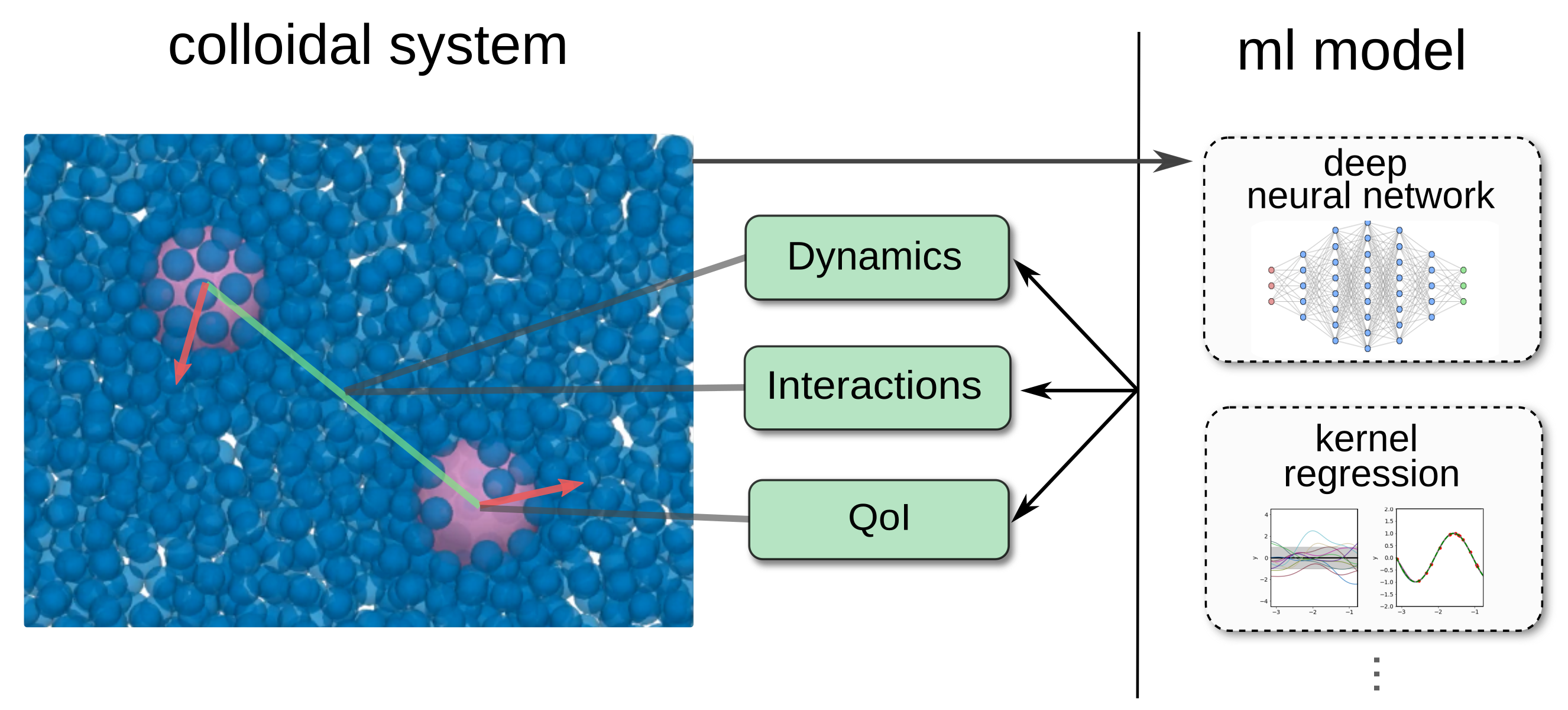}
\caption{Data-driven modeling from detailed molecular simulations can be
used to train machine learning (ML) models for performing simulations at
larger spatial-temporal scales. This can include models for the
dynamics, interactions, or for computing quantities of interest (QoI)
characterizing the system state. The colloidal system for example could
be modeled by dynamics at a larger scale with a mobility \(M\) obtained
from training. In the \texttt{MLMOD} package, the ML models can be
represented by Deep Neural Networks, Kernel Regression Models, or other
model classes.}
\end{figure}

Relative to detailed molecular dynamics simulations, this motivates a
simplified model for the effective colloid dynamics
\[\frac{d\mathbf{X}}{dt} = \mathbf{M}(\mathbf{X})\mathbf{F} 
+ k_B{T}\nabla_X \cdot \mathbf{M}(\mathbf{X}) + \mathbf{g}\]

\[< \mathbf{g}(s) \mathbf{g}(t)^T > = 2 k_B{T} \mathbf{M}(\mathbf{X}) \delta(t - s).\]
The \(\mathbf{X} \in \mathbb{R}^{3n}\) refers to the collective
configuration of all \(n\) colloids in these Smoluchowski dynamics
(\protect\hyperlink{ref-Smoluchowski:1906}{Smoluchowski, 1906}). The
\(\mathbf{g}(t)\) gives the thermal fluctuations for the temperature
corresponding to \(k_B{T}\). Here, the main objectives in this model are
to determine (i) the \emph{mobility tensor} \(M = M(\mathbf{X})\) which
captures the effective dynamic coupling between the colloidal particles,
and (ii) the \emph{interaction laws} \(\mathbf{F}\) for configurations
\(\mathbf{X}\).

Machine learning methods provide data-driven approaches for learning
representations and features for such modeling. Optimization using
appropriate loss functions and training protocols can be used to
identify system features underlying interactions, symmetries, and other
structures. In machine learning methods this is accomplished by using a
class of representations and by training with data to identify models
from this class. For making predictions in unobserved cases, this allows
for interpolation, and in some cases even extrapolation, especially when
using explicit low dimensional latent spaces or when imposing other
inductive biases (\protect\hyperlink{ref-Atzberger:2022}{Lopez \&
Atzberger, 2022}; \protect\hyperlink{ref-Atzberger:2023}{Stinis et al.,
2023}). For example, consider the colloidal example in the simplified
case when we assume the interactions can be approximated as pairwise.
The problem reduces to a model \(M = M(\mathbf{X}_1,\mathbf{X}_2)\)
depending on six dimensions. This can be further constrained to learn
only symmetric positive semi-definite tensors, for example by learning
\(L = L(\mathbf{X}_1,\mathbf{X}_2)\) to generate \(M = LL^T\).

There are many ways we can obtain the model \(M\). For example, a common
way to estimate mobility in fluid mechanics is to apply active forces
\(\mathbf{F}\) and compute the velocity response
\(< \mathbf{V} > = < {d\mathbf{X}}/{dt} > \approx \tau^{-1}< \Delta_{\tau} \mathbf{X}(t)> \approx \mathbf{M}\mathbf{F}\).
The
\(\Delta_{\tau} \mathbf{X}(t) = \mathbf{X}(t + \tau) - \mathbf{X}(t)\)
for \(\tau\) chosen carefully. For large enough forces \(\mathbf{F}\),
the thermal fluctuations can be averaged away readily by repeating this
measurement and taking the mean. In statistical mechanics, another
estimator is obtained when \(\mathbf{F} = 0\) by using the passive
fluctuations of system. A moment-based estimator commonly used is
\(M(\mathbf{X}) \approx ({2k_B{T}\tau)^{-1}} < \Delta_{\tau} \mathbf{X}(t)\Delta_{\tau} \mathbf{X}(t)^T >\)
for \(\tau\) chosen carefully. While theoretically each of these
estimators give information on \(M\), in practice there can be
subtleties such as a good choice for \(\tau\), magnitude for
\(\mathbf{F}\), and role of fluctuations. Even for these more
traditional estimators, it could still be useful for storage efficiency
and convenience to train an ML model to provide a compressed
representation and for interpolation for evaluating \(M(\mathbf{X})\).

Machine learning methods also could be used to train more directly from
simulation data for sampled colloid trajectories \(\mathbf{X}(t)\)
(\protect\hyperlink{ref-Nielsen:2000}{Nielsen et al., 2000};
\protect\hyperlink{ref-Atzberger:2023}{Stinis et al., 2023}). The
training would select an ML model \(M_\theta\) over some class of models
\(H\) parameterized by \(\theta\), such as the weights and biases of a
Deep Neural Network. For instance, this could be done by Maximum
Likelihood Estimation (MLE) or other losses from the trajectory data
\(\mathbf{X}(t)\). The MLE optimizes the objective
\[M_{\theta} = \arg\min_{M_\theta \in {H}} 
-\log\rho_\theta(\mathbf{X}(t_1),\mathbf{X}(t_2),\ldots,\mathbf{X}(t_m)).\]
The \(\rho_\theta\) denotes the likelihood probability density for the
model with \(M = M_\theta\) and observing the trajectory data
\(\{\mathbf{X}(t_i)\}\). To obtain tractable and robust training
algorithms, further approximations and regularizations may be required
to the MLE problem or alternatives used. This could include using
variational inference approaches, further restrictions on the model
architectures, priors, or other information
(\protect\hyperlink{ref-Blei:2017}{Blei et al., 2017};
\protect\hyperlink{ref-Kingma:2014}{Kingma \& Welling, 2014};
\protect\hyperlink{ref-Atzberger:2020}{Lopez \& Atzberger, 2020};
\protect\hyperlink{ref-Atzberger:2023}{Stinis et al., 2023}). Combining
such approximations with further regularizations also could help
facilitate learning, including of possible symmetries and other features
of trained models \(M(\mathbf{X}) = M_\theta\).

The \texttt{MLMOD} package provides ways for transferring such learned
models into practical simulations within LAMMPS. We discussed here one
example of a basic data-driven modeling approach for colloids. The
\texttt{MLMOD} package can be used more generally and supports broad
classes of models for incorporating machine learning results into
simulation components. Components can include the dynamics,
interactions, or computing quantities of interest. The initial
implementations we present supports the basic mobility modeling
framework as a proof-of-concept, with longer-term aims to support more
general classes of reduced dynamics and interactions in future releases.

\hypertarget{structure-of-the-package-components}{%
\section{Structure of the Package
Components}\label{structure-of-the-package-components}}

The package is organized as a module within LAMMPS that is called each
time-step and has the potential to serve multiple roles within
simulations. This includes (i) serving as a time-step integrator
updating the configuration of the system based on a specified learned
model, (ii) evaluating interactions between system components to compute
energy and forces, and (iii) computing quantities of interest (QoI) that
can be used as state information during simulations or in statistics.
The package is controlled by external XML files that specify the mode of
operation and source for pre-trained models and other information, see
the schematic in Figure 2.

\begin{figure}
\centering
\includegraphics[width=0.65\textwidth,height=\textheight]{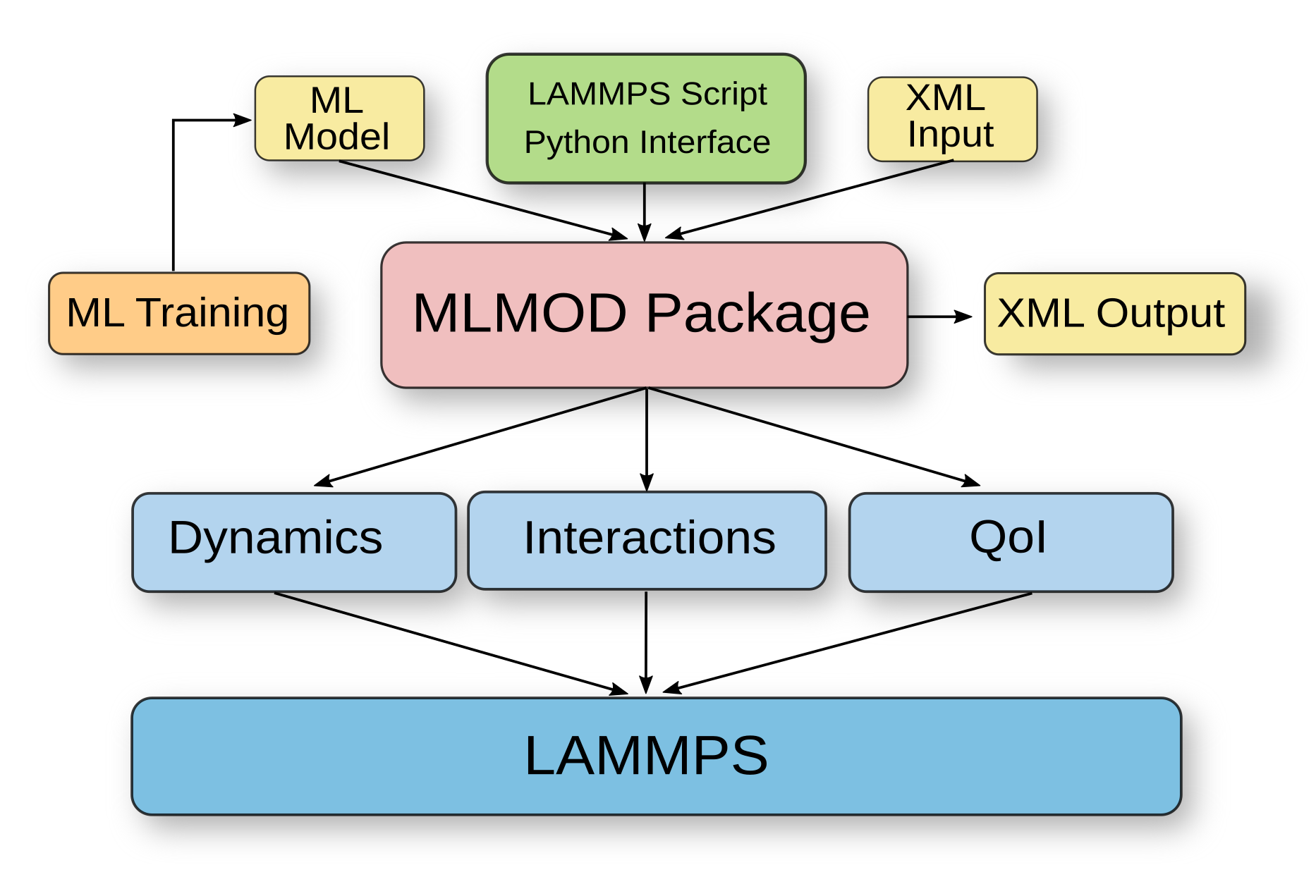}
\caption{The MLMOD Package is structured modularly with subcomponents
for providing ML models in simulations for the dynamics, interactions,
and computing quantities of interest (QoI) characterizing the system
state. The package makes use of standardized data formats such as XML
for inputs and export ML model formats from machine learning
frameworks.}
\end{figure}

The \texttt{MLMOD} Package is incorporated into a simulation by either
using the LAMMPS scripting language or the python interface. This is
done using the ``fix'' command in LAMMPS
(\protect\hyperlink{ref-Plimpton:2022}{Thompson et al., 2022}), with
this terminology historically motivated by algorithms for ``fixing''
molecular bonds as rigid each time-step. For our package the command to
set up the triggers for our algorithms is
\texttt{fix\ m1\ mlmod\ all\ filename.mlmod\_params.} This specifies the
tag ``m1'' for this fix, particle groups controlled by the package as
``all'', and the XML file of parameters. The XML file
\texttt{filename.mlmod\_params} specifies the \texttt{MLMOD} simulation
mode and where to find the associated exported ML models. An example and
more details are discussed below in the section on package usage. The
\texttt{MLMOD} Package can evaluate machine learning models using
frameworks such as C++ PyTorch API. This allows both for the possibility
of doing on-the-fly learning and for using trained models to augment
simulations.

A common approach would be to learn ML models by training on trajectory
data from detailed high fidelity simulations using a machine learning
framework, such as PyTorch (\protect\hyperlink{ref-Paszke:2019}{Paszke
et al., 2019}). Once the model is trained, it can be exported to a
portable format such as Torch
(\protect\hyperlink{ref-Collobert:2011}{Collobert et al., 2011}). The
\texttt{MLMOD} package would import these pre-trained models from Torch
files such as \texttt{trained\_model.pt}. This allows for these models
to then be invoked by \texttt{MLMOD} to provide elements for (i)
performing time-step integration to model dynamics, (ii) computing
interactions between system components, and (iii) computing quantities
of interest (QoI) for further computations or as statistics. This
provides a modular and general way for data-driven models obtained from
training with machine learning methods to be used to govern LAMMPS
simulations.

\hypertarget{example-usage-of-the-package}{%
\section{Example Usage of the
Package}\label{example-usage-of-the-package}}

We give one basic example usage of the package in the case for modeling
colloids using a mobility tensor \(M\). To set up the triggers for the
\texttt{MLMOD} package during LAMMPS simulations a typical command would
look like

\texttt{fix\ m1\ c\_group\ mlmod\ model.mlmod\_params}

The \texttt{m1} gives the tag for the fix, \texttt{c\_group} specifies
the label for the group of particles controlled by this instance of the
\texttt{MLMOD} package. The \texttt{mlmod} specifies to use the
\texttt{MLMOD} package with XML parameter file
\texttt{model.mlmod\_params}. The XML parameter file controls the
package modes and the use of associated exported ML models.

Multiple instances of \texttt{MLMOD} package are permitted and can be
used to control different groups of particles by adjusting the
\texttt{c\_group}. The package is designed with modularity so a
\emph{mode} is first defined in a parameter file and then different sets
of algorithms and parameters can be used within the same simulation. For
the mobility example, an implementation is given by the \texttt{MLMOD}
simulation mode \texttt{dX\_MF\_ML1}. For this modeling mode, a typical
parameter file would look like the following.

\begin{verbatim}
<?xml version="1.0" encoding="UTF-8"?> 
<MLMOD> 
  <model_data type="dX_MF_ML1"> 
    <M_ii_filename value="M_ii_torch.pt"/> 
    <M_ij_filename value="M_ij_torch.pt"/>
  </model_data> 
</MLMOD> 
\end{verbatim}

This specifies for an assumed mobility tensor of pairwise interactions
the models for the self-mobility responses \(M_{ii}(\mathbf{X})\) and
the pairwise mobility response
\(M_{ij}(\mathbf{X}) = M_{ji}(\mathbf{X})\), where
\(\mathbf{X} = (\mathbf{X}_{1},\mathbf{X}_{2})\). For example, a
hydrodynamic model for interactions when the two colloids of radius
\(a\) are not too close together is to use the Oseen Tensors
\(M_{ii} = (6\pi\eta a)^{-1}{I}\) and
\(M_{ij} = (8\pi\eta r)^{-1}\left({I} + r^{-2}\mathbf{r}\mathbf{r}^T \right)\).
The \(\eta\) is the fluid viscosity,
\(\mathbf{r} = \mathbf{X} _{i}(t) -\mathbf{X} _{j}(t)\) with
\(r = \|\mathbf{r}\|\) give the particle separation. The responses are
\(\mathbf{V} _{\ell} = M_{\ell m} \mathbf{F} _{m}\) with
\(\ell,m \in \{1,2\}\) and summation notation. For different
environments surrounding the colloids, these interactions would be
learned from simulation data.

The \texttt{dX\_MF\_ML1} mode indicates this type of mobility model has
interactions from learned ML models. The ML models are given by the
files \texttt{M\_ii\_torch.pt} and \texttt{M\_ij\_torch.pt}. Related
modes can also be implemented to extend models to capture more
complicated interactions or near-field effects. For example, to allow
for localized many-body interactions with ML models giving contributions
to mobility \(M(\mathbf{X})\). In this way \texttt{MLMOD} can be used
for hybrid modeling combining ML models with more traditional modeling
approaches within a unified framework.

This gives one example, the ML interactions and integrators can be more
general using any exported model from the machine learning framework.
Currently, the implementation uses PyTorch and the export format based
on torch script with \texttt{.pt} files. This allows for a variety of
models to be used ranging from those based on Deep Neural Networks,
Kernel Regression Models, and others.

\hypertarget{conclusion}{%
\section{Conclusion}\label{conclusion}}

The package \texttt{MLMOD} provides capabilities in LAMMPS for
incorporating into simulations data-driven models for dynamics and
interactions obtained from training with machine learning methods. We
describe here our initial implementation. For updates, examples, and
additional information please see \url{https://github.com/atzberg/mlmod}
and \url{http://atzberger.org/}.

\hypertarget{acknowledgements}{%
\section{Acknowledgements}\label{acknowledgements}}

Authors research supported by grants DOE Grant ASCR PHILMS DE-SC0019246,
NSF Grant DMS-1616353, and NSF Grant DMS-2306101. Authors also
acknowledge UCSB Center for Scientific Computing NSF MRSEC (DMR1121053)
and UCSB MRL NSF CNS-1725797. P.J.A. would also like to acknowledge a
hardware grant from Nvidia.

\hypertarget{references}{%
\section*{References}\label{references}}
\addcontentsline{toc}{section}{References}

\hypertarget{refs}{}
\begin{CSLReferences}{1}{0}
\leavevmode\vadjust pre{\hypertarget{ref-Abadi:2015}{}}%
Abadi, M., Agarwal, A., Barham, P., Brevdo, E., Chen, Z., Citro, C.,
Corrado, G. S., Davis, A., Dean, J., Devin, M., Ghemawat, S.,
Goodfellow, I., Harp, A., Irving, G., Isard, M., Jia, Y., Jozefowicz,
R., Kaiser, L., Kudlur, M., \ldots{} Zheng, X. (2015).
\emph{{TensorFlow}: Large-scale machine learning on heterogeneous
systems}. \url{https://www.tensorflow.org/}

\leavevmode\vadjust pre{\hypertarget{ref-Atzberger:2013}{}}%
Atzberger, Paul J. (2013). Incorporating shear into stochastic
eulerian-lagrangian methods for rheological studies of complex fluids
and soft materials. \emph{Physica D: Nonlinear Phenomena}, \emph{265},
57--70. \url{https://doi.org/10.1016/j.physd.2013.09.002}

\leavevmode\vadjust pre{\hypertarget{ref-Atzberger:2018}{}}%
Atzberger, P. J. (2018). Importance of the mathematical foundations of
machine learning methods for scientific and engineering applications.
\emph{SciML2018 Workshop, Position Paper}.
\url{https://arxiv.org/abs/1808.02213}

\leavevmode\vadjust pre{\hypertarget{ref-zenodo}{}}%
Atzberger, P. J. (2023). MLMOD package v1.0.1. \emph{Zenodo}.
\url{https://doi.org/10.5281/zenodo.8327516}

\leavevmode\vadjust pre{\hypertarget{ref-Bauer:2015}{}}%
Bauer, P., Thorpe, A., \& Brunet, G. (2015). The quiet revolution of
numerical weather prediction. \emph{Nature}, \emph{525}(7567), 47--55.
\url{https://doi.org/10.1038/nature14956}

\leavevmode\vadjust pre{\hypertarget{ref-Bird:1987}{}}%
Bird, C., R.B. (1987). \emph{Dynamics of polymeric liquids : Volume II
kinetic theory}. Wiley-Interscience.
\url{https://www.wiley.com/en-us/Dynamics+of+Polymeric+Liquids,+Volume+2:+Kinetic+Theory,+2nd+Edition-p-9780471802440}

\leavevmode\vadjust pre{\hypertarget{ref-Blei:2017}{}}%
Blei, D. M., Kucukelbir, A., \& McAuliffe, J. D. (2017). Variational
inference: A review for statisticians. \emph{Journal of the American
Statistical Association}, \emph{112}(518), 859--877.
\url{https://doi.org/10.1080/01621459.2017.1285773}

\leavevmode\vadjust pre{\hypertarget{ref-Karplus:1983}{}}%
Brooks, B. R., Bruccoleri, R. E., Olafson, B. D., States, D. J.,
Swaminathan, S. a, \& Karplus, M. (1983). CHARMM: A program for
macromolecular energy, minimization, and dynamics calculations.
\emph{Journal of Computational Chemistry}, \emph{4}(2), 187--217.
\url{https://doi.org/10.1002/jcc.540040211}

\leavevmode\vadjust pre{\hypertarget{ref-Brunton:2016}{}}%
Brunton, S. L., Proctor, J. L., \& Kutz, J. N. (2016). Discovering
governing equations from data by sparse identification of nonlinear
dynamical systems. \emph{Proceedings of the National Academy of
Sciences}, \emph{113}(15), 3932--3937.
\url{https://doi.org/10.1073/pnas.1517384113}

\leavevmode\vadjust pre{\hypertarget{ref-Collobert:2011}{}}%
Collobert, R., Kavukcuoglu, K., \& Farabet, C. (2011). Torch7: A
matlab-like environment for machine learning. \emph{BigLearn, NIPS
Workshop}. \url{https://infoscience.epfl.ch/record/192376?ln=en}

\leavevmode\vadjust pre{\hypertarget{ref-Derjaguin:1941}{}}%
Derjaguin, L., B.; Landau. (1941). Theory of the stability of strongly
charged lyophobic sols and of the adhesion of strongly charged particles
in solutions of electrolytes. \emph{Acta Physico Chemica URSS},
\emph{633}(14). \url{https://doi.org/10.1016/0079-6816(93)90013-l}

\leavevmode\vadjust pre{\hypertarget{ref-Doi:2013}{}}%
Doi, M. (2013). \emph{Soft matter physics}. Oxford University Press.
\url{https://doi.org/10.1093/acprof:oso/9780199652952.001.0001}

\leavevmode\vadjust pre{\hypertarget{ref-Giessen:2020}{}}%
Giessen, E. van der, Schultz, P. A., Bertin, N., Bulatov, V. V., Cai,
W., Csányi, G., Foiles, S. M., Geers, M. G. D., González, C., Hütter,
M., Kim, W. K., Kochmann, D. M., LLorca, J., Mattsson, A. E., Rottler,
J., Shluger, A., Sills, R. B., Steinbach, I., Strachan, A., \& Tadmor,
E. B. (2020). Roadmap on multiscale materials modeling. \emph{Modelling
and Simulation in Materials Science and Engineering}, \emph{28}(4),
043001. \url{https://doi.org/10.1088/1361-651x/ab7150}

\leavevmode\vadjust pre{\hypertarget{ref-Goodfellow:2016}{}}%
Goodfellow, I., Bengio, Y., \& Courville, A. (2016).
\emph{\href{https://www.deeplearningbook.org/}{Deep learning}}. The MIT
Press. ISBN:~0262035618

\leavevmode\vadjust pre{\hypertarget{ref-Hastie:2001}{}}%
Hastie, T., Tibshirani, R., \& Friedman, J. (2001). \emph{Elements of
statistical learning}. Springer New York Inc.
\url{https://doi.org/10.1007/978-0-387-84858-7}

\leavevmode\vadjust pre{\hypertarget{ref-Hinton:2006}{}}%
Hinton, G., \& Salakhutdinov, R. (2006). Reducing the dimensionality of
data with neural networks. \emph{Science}, \emph{313}(5786), 504--507.
\url{https://doi.org/10.1126/science.1127647}

\leavevmode\vadjust pre{\hypertarget{ref-Jones:2002}{}}%
Jones, R. A. L., Jones, R. A. L., \& R Jones, P. (2002).
\emph{\href{https://books.google.com/books?id=Hl/_HBPUvoNsC}{Soft
condensed matter}}. OUP Oxford. ISBN:~9780198505891

\leavevmode\vadjust pre{\hypertarget{ref-Karplus:2002}{}}%
Karplus, M., \& McCammon, J. A. (2002). Molecular dynamics simulations
of biomolecules. \emph{Nature Structural Biology}, \emph{9}(9),
646--652. \url{https://doi.org/10.1038/nsb0902-646}

\leavevmode\vadjust pre{\hypertarget{ref-Kimura:2009}{}}%
Kimura, Y. (2009). Microrheology of soft matter. \emph{J. Phys. Soc.
Jpn.}, \emph{78}(4), 8--8. \url{https://doi.org/10.1143/JPSJ.78.041005}

\leavevmode\vadjust pre{\hypertarget{ref-Kingma:2014}{}}%
Kingma, D. P., \& Welling, M. (2014). Auto-encoding variational bayes.
\emph{2nd International Conference on Learning Representations, {ICLR}
2014, Banff, AB, Canada, April 14-16, 2014, Conference Track
Proceedings}. \url{http://arxiv.org/abs/1312.6114}

\leavevmode\vadjust pre{\hypertarget{ref-Atzberger:2020}{}}%
Lopez, R., \& Atzberger, P. J. (2020). \emph{Variational autoencoders
for learning nonlinear dynamics of physical systems}.
\url{https://arxiv.org/abs/2012.03448}

\leavevmode\vadjust pre{\hypertarget{ref-Atzberger:2022}{}}%
Lopez, R., \& Atzberger, P. J. (2022). GD-VAEs: Geometric dynamic
variational autoencoders for learning nonlinear dynamics and dimension
reductions. \emph{arXiv Preprint arXiv:2206.05183}.
\url{https://arxiv.org/abs/2206.05183}

\leavevmode\vadjust pre{\hypertarget{ref-Lubensky:1997}{}}%
Lubensky, T. C. (1997). Soft condensed matter physics. \emph{Solid State
Communications}, \emph{102}(2-3), 187-197-.
\url{https://doi.org/10.1016/S0038-1098(96)00718-1}

\leavevmode\vadjust pre{\hypertarget{ref-Lusk:2011}{}}%
Lusk, M. T., \& Mattsson, A. E. (2011). High-performance computing for
materials design to advance energy science. \emph{MRS Bulletin},
\emph{36}(3), 169--174. \url{https://doi.org/10.1557/mrs.2011.30}

\leavevmode\vadjust pre{\hypertarget{ref-Mccammon:1988}{}}%
McCammon, J. A., \& Harvey, S. C. (1988). \emph{Dynamics of proteins and
nucleic acids}. Cambridge University Press.
\url{https://doi.org/10.1017/CBO9781139167864}

\leavevmode\vadjust pre{\hypertarget{ref-Murr:2016}{}}%
Murr, L. E. (2016). Computer simulations in materials science and
engineering. In \emph{Handbook of materials structures, properties,
processing and performance} (pp. 1--15). Springer International
Publishing. \url{https://doi.org/10.1007/978-3-319-01905-5_60-2}

\leavevmode\vadjust pre{\hypertarget{ref-Nielsen:2000}{}}%
Nielsen, J. N., Madsen, H., \& Young, P. C. (2000). Parameter estimation
in stochastic differential equations: An overview. \emph{Annual Reviews
in Control}, \emph{24}, 83--94.
\url{https://doi.org/10.1016/S1367-5788(00)90017-8}

\leavevmode\vadjust pre{\hypertarget{ref-Pan:2021}{}}%
Pan, J. (2021). Scaling up system size in materials simulation.
\emph{Nature Computational Science}, \emph{1}(2), 95--95.
\url{https://doi.org/10.1038/s43588-021-00034-x}

\leavevmode\vadjust pre{\hypertarget{ref-Paszke:2019}{}}%
Paszke, A., Gross, S., Massa, F., Lerer, A., Bradbury, J., Chanan, G.,
Killeen, T., Lin, Z., Gimelshein, N., Antiga, L., Desmaison, A., Kopf,
A., Yang, E., DeVito, Z., Raison, M., Tejani, A., Chilamkurthy, S.,
Steiner, B., Fang, L., \ldots{} Chintala, S. (2019). PyTorch: An
imperative style, high-performance deep learning library. In H. Wallach,
H. Larochelle, A. Beygelzimer, F. dAlché-Buc, E. Fox, \& R. Garnett
(Eds.), \emph{Advances in neural information processing systems 32} (pp.
8024--8035). Curran Associates, Inc.
\href{http://papers.neurips.cc/paper/9015-pytorch-an-imperative-\%0A\%20\%20\%20\%20\%20\%20\%20\%20\%20\%20\%20\%20\%20\%20\%20\%20\%20\%20style-high-performance-deep-learning-library.pdf}{http://papers.neurips.cc/paper/9015-pytorch-an-imperative-
style-high-performance-deep-learning-library.pdf}

\leavevmode\vadjust pre{\hypertarget{ref-Rasmussen:2004}{}}%
Rasmussen, C. E. (2004). Gaussian processes in machine learning. In O.
Bousquet, U. von Luxburg, \& G. Rätsch (Eds.), \emph{Advanced lectures
on machine learning: ML summer schools 2003, canberra, australia,
february 2 - 14, 2003, t{ü}bingen, germany, august 4 - 16, 2003, revised
lectures} (pp. 63--71). Springer Berlin Heidelberg.
\url{https://doi.org/10.1007/978-3-540-28650-9_4}

\leavevmode\vadjust pre{\hypertarget{ref-Richardson:2007}{}}%
Richardson, L. F. (2007). \emph{Weather prediction by numerical
process}. Cambridge university press.
\url{https://archive.org/details/weatherpredictio00richrich}

\leavevmode\vadjust pre{\hypertarget{ref-Sanbonmatsu:2007}{}}%
Sanbonmatsu, K., \& Tung, C.-S. (2007). High performance computing in
biology: Multimillion atom simulations of nanoscale systems.
\emph{Journal of Structural Biology}, \emph{157}(3), 470--480.
\url{https://doi.org/10.1016/j.jsb.2006.10.023}

\leavevmode\vadjust pre{\hypertarget{ref-Schmidt:2009}{}}%
Schmidt, M., \& Lipson, H. (2009). \emph{Distilling free-form natural
laws from experimental data}. \emph{324}, 81--85.
\url{https://doi.org/10.1126/science.1165893}

\leavevmode\vadjust pre{\hypertarget{ref-Scholkopf:2001}{}}%
Scholkopf, B., \& Smola, A. J. (2001).
\emph{\href{https://vdoc.mx/documents/learning-with-kernels-support-vector-machines-regularization-optimization-and-beyond-14ipl8tqmq3o\#}{Learning
with kernels: Support vector machines, regularization, optimization, and
beyond}}. MIT Press. ISBN:~0262194759

\leavevmode\vadjust pre{\hypertarget{ref-Atzberger:2018b}{}}%
Sidhu, I., Frischknecht, A. L., \& Atzberger, P. J. (2018).
Electrostatics of nanoparticle-wall interactions within nanochannels:
Role of double-layer structure and ion-ion correlations. \emph{ACS
Omega}, \emph{3}(9), 11340--11353.
\url{https://doi.org/10.1021/acsomega.8b01393}

\leavevmode\vadjust pre{\hypertarget{ref-Smoluchowski:1906}{}}%
Smoluchowski, V. (1906). Drei vorträge über diffusion, brownsche
molekularbewegung und koagulation von kolloidteilchen. \emph{Ann. Phys},
\emph{21}, 756.
\url{https://jbc.bj.uj.edu.pl/Content/387533/PDF/FIZART_SMOLUCHOWSKI_00093.pdf}

\leavevmode\vadjust pre{\hypertarget{ref-Atzberger:2023}{}}%
Stinis, P., Daskalakis, C., \& Atzberger, P. J. (2023). SDYN-GANs:
Adversarial learning methods for multistep generative models for general
order stochastic dynamics. \emph{arXiv Preprint arXiv:2302.03663}.
\url{https://arxiv.org/abs/2302.03663}

\leavevmode\vadjust pre{\hypertarget{ref-Plimpton:2022}{}}%
Thompson, A. P., Aktulga, H. M., Berger, R., Bolintineanu, D. S., Brown,
W. M., Crozier, P. S., in 't Veld, P. J., Kohlmeyer, A., Moore, S. G.,
Nguyen, T. D., Shan, R., Stevens, M. J., Tranchida, J., Trott, C., \&
Plimpton, S. J. (2022). LAMMPS - a flexible simulation tool for
particle-based materials modeling at the atomic, meso, and continuum
scales. \emph{Computer Physics Communications}, \emph{271}, 108171.
\url{https://doi.org/10.1016/j.cpc.2021.108171}

\leavevmode\vadjust pre{\hypertarget{ref-Washington:2009}{}}%
Washington, W. M., Buja, L., \& Craig, A. (2009). The computational
future for climate and earth system models: On the path to petaflop and
beyond. \emph{Philosophical Transactions of the Royal Society A:
Mathematical, Physical and Engineering Sciences}, \emph{367}(1890),
833--846. \url{https://doi.org/10.1098/rsta.2008.0219}

\end{CSLReferences}

\end{document}